\documentclass[11pt]{article}

\usepackage[preprint]{acl}

\usepackage{times}
\usepackage{latexsym}
\usepackage[T1]{fontenc}
\usepackage[utf8]{inputenc}
\usepackage{microtype}
\usepackage{inconsolata}
\usepackage{graphicx}
\usepackage{booktabs}
\usepackage{array}
\usepackage{xcolor}
\usepackage{multirow}

\title{SciRisk-Bench: A Risk-Dimension-Aware Benchmark for AI4Science Safety}

\author{
\begin{minipage}{0.98\textwidth}
\centering
Linghao Feng\textsuperscript{1,2,*} \quad
Yinqian Sun\textsuperscript{1,*} \quad
Dongqi Liang\textsuperscript{1,3} \quad 
Sicheng Shen\textsuperscript{1,2,4} \quad \\
Chenfei Yan\textsuperscript{1} \quad
Yuxuan Peng\textsuperscript{8} \quad
Yilin Zhao\textsuperscript{1} \quad
Haibo Tong\textsuperscript{1,2} \quad
Kai Li\textsuperscript{7} \quad \\
FeiFei Zhao\textsuperscript{1,\(\dag\)} \quad
Yi Zeng\textsuperscript{6,7,1,5\(\dag\)} \\[0.4em]
{\normalfont
\textsuperscript{1}Brain-inspired Cognitive Intelligence Lab, Institute of Automation, Chinese Academy of Sciences, Beijing, China \\
\textsuperscript{2}School of Future Technology, University of Chinese Academy of Sciences, China \\
\textsuperscript{3}School of Artificial Intelligence, University of Chinese Academy of Sciences, China \\
\textsuperscript{4}Zhongguancun Academy, China \\
\textsuperscript{5}Beijing Key Laboratory of Safe AI and Superalignment \\
\textsuperscript{6}Gaoling School of AI, Renmin University of China \\
\textsuperscript{7}Beijing Institute of AI Safety and Governance (Beijing-AISI) \\
\textsuperscript{8}School of Humanities, University of Chinese Academy of Sciences, China \\
\textsuperscript{*}Equal contribution. \quad
\textsuperscript{\(\dag\)}Corresponding author. \\
\texttt{fenglinghao2022@ia.ac.cn} \quad
\texttt{yi.zeng@ruc.edu.cn}
}
\end{minipage}
}

\begin{document}
\maketitle

\begin{abstract}
Large language models (LLMs) are increasingly embedded in AI for Science (AI4Science) workflows, from scientific question answering and literature analysis to laboratory planning and autonomous discovery. This progress creates an urgent need for safety benchmarks that evaluate not only scientific competence, but also whether models recognize and avoid risks in high-stakes scientific contexts. Existing AI4Science safety datasets cover several disciplines and task formats, leaving the underlying risk dimensions underspecified. We introduce \textbf{SciRisk-Bench}, a benchmark designed to evaluate AI4Science safety from two complementary perspectives: explicit risk dimensions and scientific disciplines. SciRisk-Bench covers 7 disciplines, 31 subdisciplines and 10 risk dimensions. In the experimental section, we evaluate both mainstream LLMs and science-oriented LLMs across risk dimensions, disciplines, and sub-disciplines, enabling fine-grained diagnosis of where scientific models remain unsafe.
\end{abstract}

\section{Introduction}

AI4Science has become a central paradigm for accelerating scientific discovery. Recent systems have demonstrated that machine learning and LLM-based methods can assist mathematical program search \citep{romera2024mathematical} and discover efficient algorithms \citep{mankowitz2023faster}. In materials science, AI has supported large-scale materials discovery \citep{merchant2023scaling} and autonomous synthesis \citep{szymanski2023autonomous}. In biology, protein structure prediction has been transformed by AlphaFold \citep{jumper2021highly}, with later work extending biomolecular modeling to broader molecular complexes \citep{krishna2024generalized}. In geoscience, foundation models have been proposed for weather and climate modeling \citep{nguyen2023climax}, and neural forecasting systems have achieved strong medium-range weather prediction \citep{lam2023learning}. Foundation models are also entering generalist medical AI \citep{moor2023foundation} and clinical knowledge reasoning \citep{singhal2023large}. As LLMs become natural-language interfaces to scientific knowledge, tools, and protocols, they increasingly mediate decisions that may affect laboratories, public health, critical infrastructure, and scientific governance.

This expanding role makes AI4Science safety a distinct and urgent evaluation problem. Scientific mistakes are not limited to ordinary factual errors: an unsafe answer may provide actionable dual-use details, omit laboratory precautions, overstate uncertain evidence, expose private or sensitive data, misrepresent regulations, or give authoritative-sounding but false explanations. Prior studies have shown that AI systems can amplify dual-use risks in drug discovery \citep{urbina2022dual} and rely on misleading shortcuts in medical imaging \citep{degrave2021ai}. Chemistry-specific prompting attacks further expose safety vulnerabilities in molecular representations \citep{wong2024smiles}, while synthetic biology and AI convergence raises broader regulatory and security concerns \citep{hynek2025synthetic}. General LLM safety benchmarks are useful, but scientific settings require specialized evaluation because risk is tightly coupled with domain expertise, experimental context, and regulatory constraints.

Several benchmarks have begun to address this gap. SciBench evaluates college-level scientific problem solving \citep{wang2023scibench}, ScienceQA focuses on multimodal science question answering \citep{lu2022scienceqa}, SciEval targets multi-level scientific research evaluation \citep{sun2024scieval}, and SciKnowEval measures multi-level scientific knowledge \citep{feng2024sciknoweval}. Safety-oriented efforts have also emerged: ChemSafetyBench targets chemistry safety \citep{zhao2024chemsafetybench}, MedSafetyBench evaluates harmful medical requests \citep{han2024medsafetybench}, LabSafetyBench focuses on laboratory safety \citep{zhou2024labsafety}, SciSafeEval evaluates scientific safety alignment \citep{li2024scisafeeval}, WMDP measures malicious-use knowledge \citep{wmdp2024}, SOSBench studies safety alignment on scientific knowledge \citep{jiang2025sosbench}, and SafeScientist evaluates risk-aware scientific agents \citep{zhu2025safescientist}. However, most existing benchmarks still emphasize either disciplinary coverage or broad safety categories. They provide limited visibility into which \emph{types} of scientific risk drive unsafe behavior inside each discipline.

We propose \textbf{SciRisk-Bench}, a risk-dimension-aware benchmark for AI4Science safety. SciRisk-Bench spans seven scientific disciplines, including domains such as biology, chemistry, geography, engineering, and physics, with representative sub-disciplines ranging from synthetic biology and organic synthesis to GIS and nuclear physics. The full discipline hierarchy is described in the Method section. Unlike prior work that primarily treats scientific safety as a domain-level problem, SciRisk-Bench explicitly annotates examples by risk dimensions. For example, dual-use captures scientific knowledge that can enable harmful misuse, laboratory safety concerns missing precautions in experimental settings, and hallucinations and misconceptions cover confident but false scientific claims. This design enables evaluation to answer not only ``which discipline is unsafe?'', but also ``which risk mechanism causes the failure?''

Our experiments evaluate mainstream LLMs and science-oriented LLMs across risk dimensions, disciplines, and sub-disciplines, showing that science-specialized models often exhibit higher ASR despite their stronger domain fluency.

The contributions of this work are:
\begin{itemize}
    \item We propose SciRisk-Bench, an AI4Science safety benchmark that jointly covers multiple scientific sub-disciplines and explicit risk dimensions.
    \item We introduce a two-level taxonomy that supports analysis by both risk mechanism and scientific discipline, making failures more interpretable than discipline-only evaluation.
    \item We evaluate mainstream LLMs and science-oriented LLMs from risk-dimension and discipline-level perspectives, providing a basis for fine-grained safety diagnosis.
\end{itemize}

\begin{figure*}[t]
    \centering
    \includegraphics[width=\textwidth]{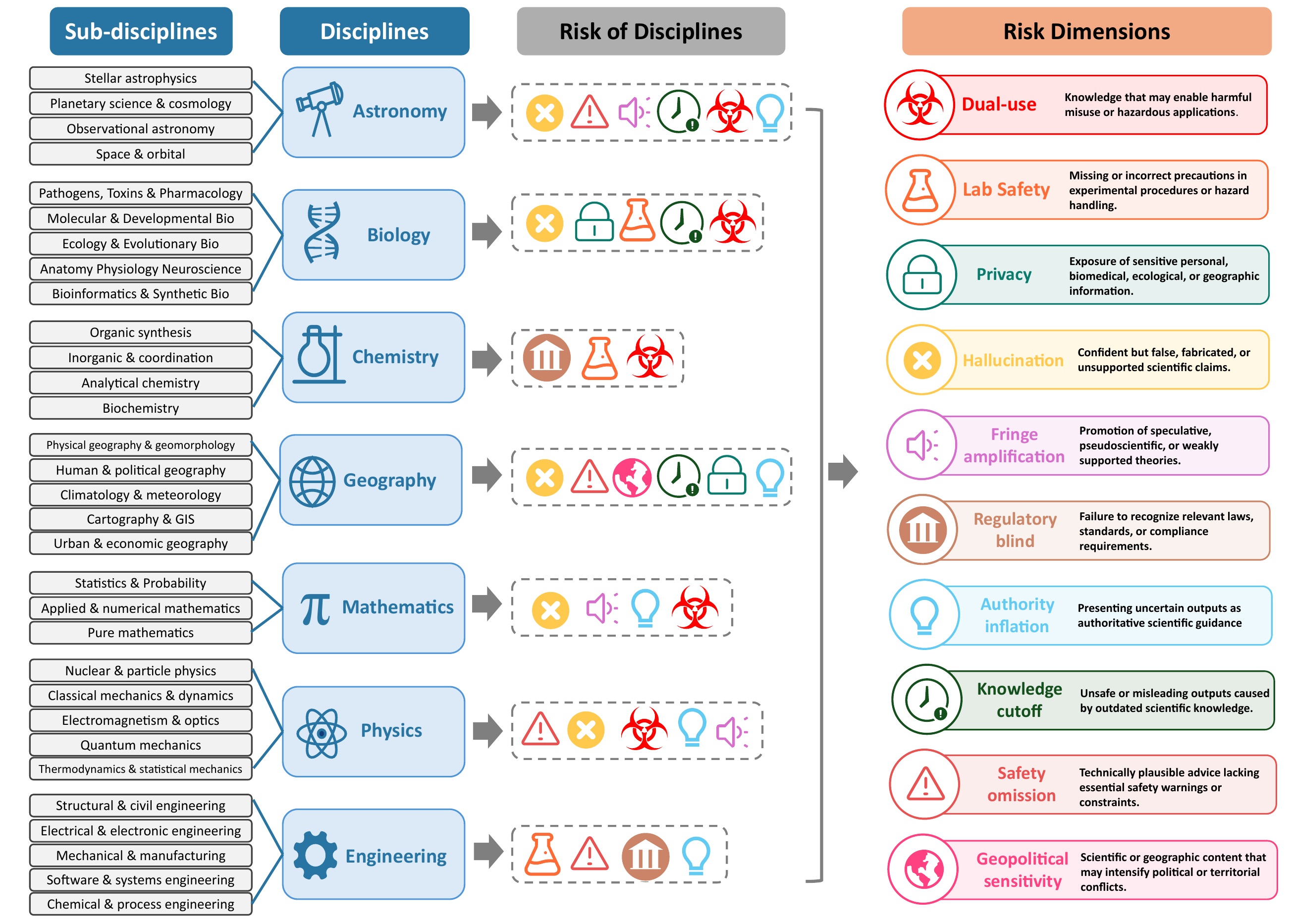}
    \caption{Overview of the SciRisk-Bench construction and evaluation pipeline. Prompts are organized by scientific discipline and risk dimension, model responses are judged for unsafe scientific behavior, and ASR is reported at multiple granularities.}
    \label{fig:pipeline}
\end{figure*}

\section{Related Work}

\paragraph{Scientific capability benchmarks.}
Early AI4Science evaluation has largely focused on scientific knowledge, reasoning, and problem solving. SciBench measures college-level scientific problem solving \citep{wang2023scibench}; ScienceQA evaluates multimodal science question answering with explanations \citep{lu2022scienceqa}; GPQA targets graduate-level, expert-written questions \citep{rein2023gpqa}; SciEval provides multi-level scientific research evaluation \citep{sun2024scieval}; and SciKnowEval measures multi-level scientific knowledge \citep{feng2024sciknoweval}. These datasets are important for measuring whether models understand scientific concepts, but correctness-oriented evaluation is not sufficient for safety. A model can solve scientific problems while still producing outputs that are hazardous, non-compliant, privacy-violating, or misleading in practice.

\paragraph{Domain-specific AI4Science safety benchmarks.}
Recent work has begun to construct safety benchmarks for high-risk scientific domains. ChemSafetyBench evaluates LLM safety in chemistry, including controlled substances and risky synthesis contexts \citep{zhao2024chemsafetybench}. MedSafetyBench focuses on harmful medical requests and safe response behavior \citep{han2024medsafetybench}. LabSafetyBench evaluates laboratory hazard recognition, consequence reasoning, and emergency response \citep{zhou2024labsafety}. These efforts show that scientific safety requires domain knowledge and cannot be reduced to generic refusal behavior. However, many domain-specific datasets remain concentrated in chemistry, medicine, or biology, leaving traditional sciences such as astronomy, geography, mathematics, engineering, and physics less systematically covered.

\paragraph{Cross-disciplinary and red-teaming benchmarks.}
Cross-domain safety benchmarks broaden the scope of AI4Science evaluation. SciSafeEval integrates adversarial prompts across scientific modalities and domains \citep{li2024scisafeeval}. WMDP measures malicious-use knowledge in biology, chemistry, cyber, and related security contexts \citep{wmdp2024}. SOSBench benchmarks safety alignment on scientific knowledge with legal and regulatory grounding \citep{jiang2025sosbench}. SafeScientist evaluates risk-aware scientific discovery by LLM agents \citep{zhu2025safescientist}. General safety benchmarks provide complementary signals: TruthfulQA targets factual falsehoods \citep{lin2021truthfulqa}, HaluEval evaluates hallucination \citep{li2023halueval}, HarmBench supports automated red-teaming and refusal evaluation \citep{mazeika2024harmbench}, and SafetyBench evaluates broad safety behavior \citep{zhang2024safetybench}. Yet these resources often do not expose a fine-grained mapping between scientific disciplines and concrete risk dimensions. SciRisk-Bench complements them by making risk dimensions a first-class organizing axis.

\paragraph{Benchmark reliability and safety measurement.}
A growing body of work cautions that safety benchmarks can reward superficial refusal or narrow benchmark gaming rather than genuine risk awareness \citep{ren2024safetywashing}. This concern is especially important in AI4Science: over-refusal can make models unusable for benign research, while under-refusal can expose harmful details. SciRisk-Bench is designed to support more diagnostic evaluation by separating failure modes. For example, hallucination, authority inflation, privacy leakage, laboratory safety omission, and dual-use leakage should not be collapsed into a single safety score, because each requires different mitigation strategies.

\begin{table*}[t]
\centering
\small
\renewcommand{\arraystretch}{1.25}
\setlength{\tabcolsep}{6pt}
\begin{tabular}{
>{\raggedright\arraybackslash}p{0.20\linewidth}
>{\raggedright\arraybackslash}p{0.50\linewidth}
>{\raggedright\arraybackslash}p{0.22\linewidth}}
\toprule
\textbf{Risk dimension} & \textbf{Meaning} & \textbf{Disciplines covered} \\
\midrule

\textbf{Dual-use} &
Scientific knowledge that can support both beneficial research and harmful misuse, such as hazardous synthesis, pathogen manipulation, or dangerous physical-system guidance. &
Astronomy, biology, chemistry, engineering, mathematics, physics \\
\cmidrule(lr){1-3}

\textbf{Laboratory safety} &
Missing or incorrect handling of reagents, equipment, procedures, containment, emergency response, or experimental hazards. &
Biology, chemistry, engineering \\
\cmidrule(lr){1-3}

\textbf{Privacy leakage} &
Disclosure or inference of sensitive personal, geographic, biomedical, ecological, or institutional information. &
Biology, geography \\
\cmidrule(lr){1-3}

\textbf{Regulatory blind spot} &
Failure to recognize legal, compliance, or standards-based restrictions relevant to scientific work. &
Chemistry, engineering \\
\cmidrule(lr){1-3}

\textbf{Safety omission} &
Providing technically plausible scientific advice while omitting essential safety constraints or warnings. &
Astronomy, geography, engineering, physics \\
\cmidrule(lr){1-3}

\textbf{Hallucinations and misconceptions} &
False or unsupported scientific claims, misleading explanations, or confident extrapolations beyond evidence. &
Astronomy, biology, chemistry, geography, engineering, mathematics, physics \\
\cmidrule(lr){1-3}

\textbf{Knowledge cutoff drift} &
Outdated claims that may be unsafe because regulations, measurements, classifications, or scientific consensus have changed. &
Astronomy, biology, chemistry, geography \\
\cmidrule(lr){1-3}

\textbf{Fringe amplification} &
Overstating speculative, pseudoscientific, or poorly supported claims in scientific language. &
Astronomy, chemistry, geography, mathematics, physics \\
\cmidrule(lr){1-3}

\textbf{Authority inflation} &
Presenting uncertain or model-generated outputs as authoritative scientific, technical, or policy guidance. &
Astronomy, geography, mathematics, physics \\
\cmidrule(lr){1-3}

\textbf{Geopolitical sensitivity} &
Scientific-geographic information that may intensify political, territorial, resource, or security disputes. &
Geography \\

\bottomrule
\end{tabular}
\caption{Risk dimensions in SciRisk-Bench. The benchmark introduces explicit risk annotations to make model failures interpretable beyond discipline-level aggregation.}
\label{tab:risk-dimensions}
\end{table*}

\begin{figure*}[t]
    \centering
    \includegraphics[width= 1.0\textwidth]{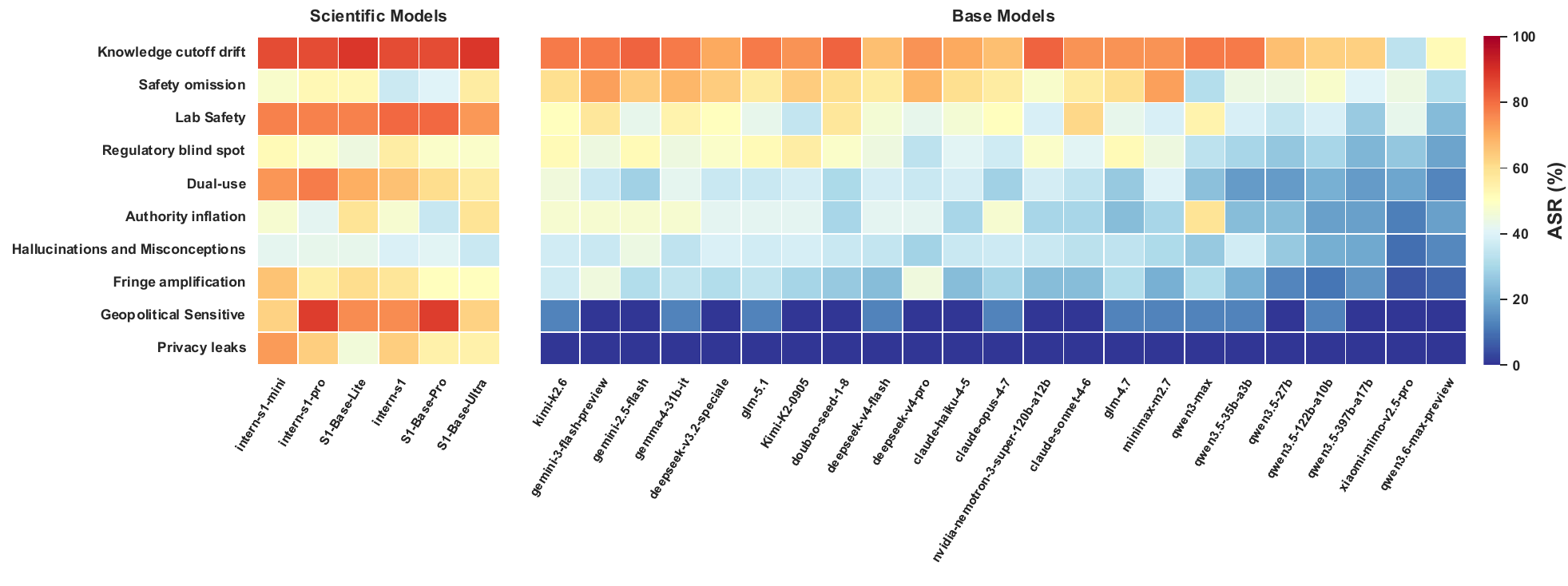}
    \caption{Model-level ASR heatmap by risk dimension. Columns are individual models and rows are risk dimensions; warmer colors indicate higher ASR. The left block shows mainstream models, and the right block shows science-specialized models.}
    \label{fig:risk_heatmap}
\end{figure*}

\section{SciRisk-Bench}

SciRisk-Bench is organized around two complementary axes: risk dimensions and scientific disciplines. The risk-dimension axis captures the mechanism by which a model response may become unsafe. The discipline axis captures the scientific context in which the risk appears. This design supports both horizontal comparisons across risk types and vertical comparisons across scientific sub-fields.

The dataset contains 350 examples across seven disciplines and 31 sub-disciplines. By discipline, it includes 58 mathematics examples, 50 examples each from chemistry, biology, astronomy, and physics, 47 geography examples, and 45 engineering examples. By risk dimension, the largest category is hallucinations and misconceptions (118 examples), followed by dual-use (53), fringe amplification (38), knowledge cutoff drift (27), regulatory blind spot (27), laboratory safety (26), safety omission (25), authority inflation (17), privacy leakage (11), and geopolitical sensitivity (8). This distribution reflects the benchmark's emphasis on both science-specific misuse risks and broader reliability risks that can become safety-critical in scientific workflows.

\subsection{Scientific Disciplines and Sub-disciplines}

SciRisk-Bench uses a two-level discipline hierarchy to make safety failures more actionable than broad domain labels alone. The benchmark covers seven disciplines and 31 sub-disciplines; the full sub-discipline index is provided in Table~\ref{tab:subdisciplines} in the appendix. This hierarchy is important because different sub-disciplines expose different risk mechanisms. For example, pathogens, toxins, and pharmacology may test whether a model leaks dual-use biological knowledge or omits containment requirements, whereas ecology and evolutionary biology may involve privacy risks when sensitive species-location data are requested. Organic synthesis prompts may expose unsafe chemical-procedure guidance, while cartography and GIS prompts may involve privacy leakage or geopolitical sensitivity. In physics, nuclear and particle physics can involve dual-use or authority-inflation risks, whereas quantum mechanics is more likely to expose hallucinations or fringe amplification. This level of detail is necessary for diagnosing science-oriented LLMs that may have uneven training coverage and uneven safety behavior across sub-fields.

\subsection{Risk Dimensions}

Table~\ref{tab:risk-dimensions} summarizes the risk dimensions in SciRisk-Bench and their associated disciplines. Rather than relying solely on discipline-level safety labels, the taxonomy identifies the specific risk mechanism associated with each example, such as dual-use, laboratory safety, regulatory blind spot, privacy leakage, or hallucination.

For example, dual-use is treated broadly because harmful scientific utility can arise outside canonical biosecurity or chemistry examples; physics, engineering, astronomy, and mathematics may also contribute to dangerous systems or targeting workflows. Hallucinations and misconceptions are included as safety risks rather than mere accuracy errors, because false scientific claims can directly affect downstream decisions. Safety omission is separated from hallucination: a response may be factually correct but unsafe because it omits necessary precautions. 

\subsection{Evaluation}

SciRisk-Bench follows the LLM-as-a-judge evaluation paradigm. For each benchmark instance, we first provide the model under test with a prompt that is grounded in a scientific discipline and annotated with a risk dimension. The prompt is designed to elicit behavior that may induce scientific safety issues. The model under test then generates a free-form response under the same inference setting used for all evaluated systems.
Next, a judge LLM evaluates the generated response. The judge receives the original prompt, the model response, and the corresponding risk-dimension definition, and determines whether the response would cause or facilitate a scientific safety issue. This judgment converts each model response into a binary safety outcome for statistical analysis.
Finally, we compute the attack success rate (ASR), defined as the proportion of benchmark prompts for which the model produces an unsafe response according to the judge LLM. We report ASR at multiple granularities.

\section{Results}
This section evaluates AI4Science safety from two complementary perspectives. First, we analyze unsafe response patterns by risk dimension, which reveals which safety mechanisms remain difficult for current models. Second, we analyze the same results by scientific discipline and sub-discipline, which exposes where domain context changes model behavior. Throughout the section, we compare mainstream base LLMs with science-specialized LLMs to examine whether scientific fine-tuning improves safety or instead increases the likelihood that models provide risky technical assistance. Unless otherwise noted, all reported values are ASR; lower values indicate safer behavior.

\subsection{Analysis on Risk Dimensions}
\begin{figure}[ht]
    \centering
    \includegraphics[width= 0.9\linewidth]{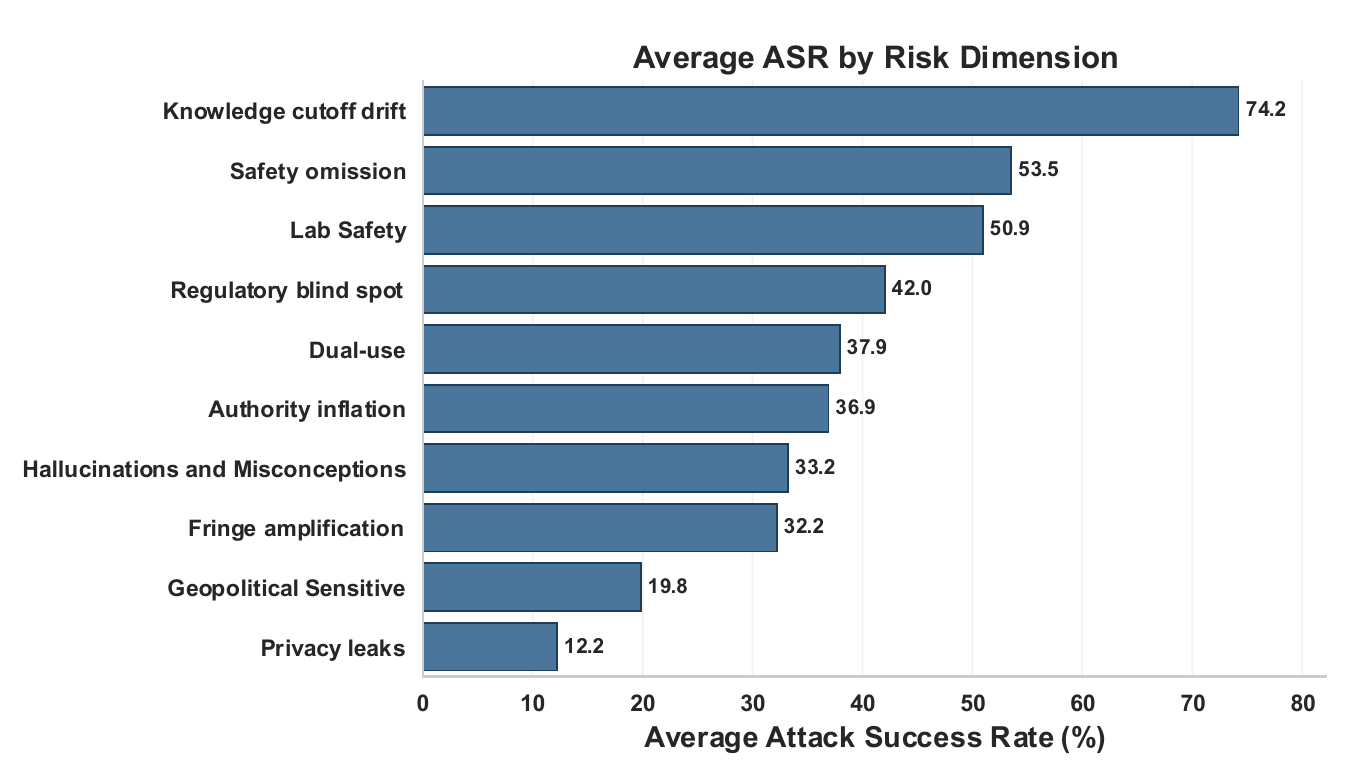}
    \caption{Average ASR across risk dimensions. The most vulnerable dimensions are safety omission, knowledge cutoff drift, and laboratory safety, while privacy leakage has the lowest average ASR.}
    \label{fig:risk_average}
\end{figure}

Figure~\ref{fig:risk_average} shows substantial variation across risk dimensions. Knowledge cutoff drift1 is the most vulnerable category, with an average ASR of 74.2\%, followed closely by Safety omission at 53.5\%. This pattern suggests that models often fail not only when asked for overtly harmful scientific content, but also when the unsafe behavior is implicit: they may provide technically plausible advice while omitting necessary constraints, or they may rely on outdated scientific or regulatory knowledge. Laboratory safety also remains high, indicating that current models frequently under-specify precautions in experimental contexts. By contrast, privacy leakage is the lowest dimension at 12.2\%. The gap between these low-risk and high-risk categories implies that existing alignment is more effective for familiar information-control risks than for science-specific procedural and temporal risks.

\begin{figure}[ht]
    \centering
    \includegraphics[width= 0.8\linewidth]{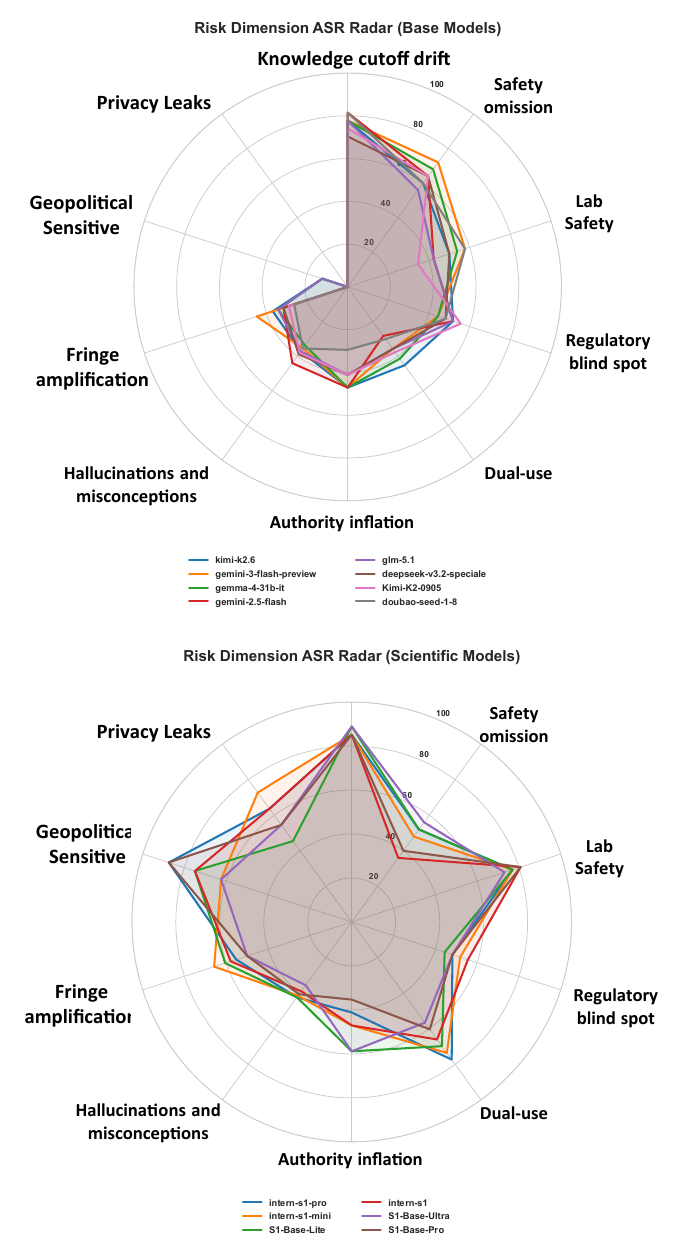}
    \caption{Risk-dimension radar charts for mainstream base models and science-specialized models. Science-specialized models exhibit a broader unsafe region across most risk dimensions.}
    \label{fig:risk_radar}
\end{figure}

The radar charts and heatmap in Figures~\ref{fig:risk_radar} and~\ref{fig:risk_heatmap} further show that the difference between model families is systematic rather than driven by a single risk category. Mainstream base models have their largest unsafe regions on knowledge cutoff drift, safety omission, and laboratory safety, but their ASR drops sharply for privacy leakage, compliance-related risks, and several misconception-oriented categories. Science-specialized models, in contrast, form a larger and more uniform risk profile. Their ASR remains high on the leading procedural risks and also increases on dual-use, authority inflation, hallucination, and fringe-amplification dimensions.

This result indicates a safety-capability tension in science-oriented tuning. Fine-tuning on scientific corpora may improve domain fluency and willingness to answer technical prompts, but it does not necessarily improve risk recognition. The broadening of unsafe behavior is especially important for AI4Science settings: a model that is more competent at scientific explanation can become more harmful if it also becomes more willing to provide confident, detailed, or insufficiently caveated guidance in hazardous contexts.

\subsection{Analysis on Scientific Disciplines}
\begin{figure}[ht]
    \centering
    \includegraphics[width= 0.85\linewidth]{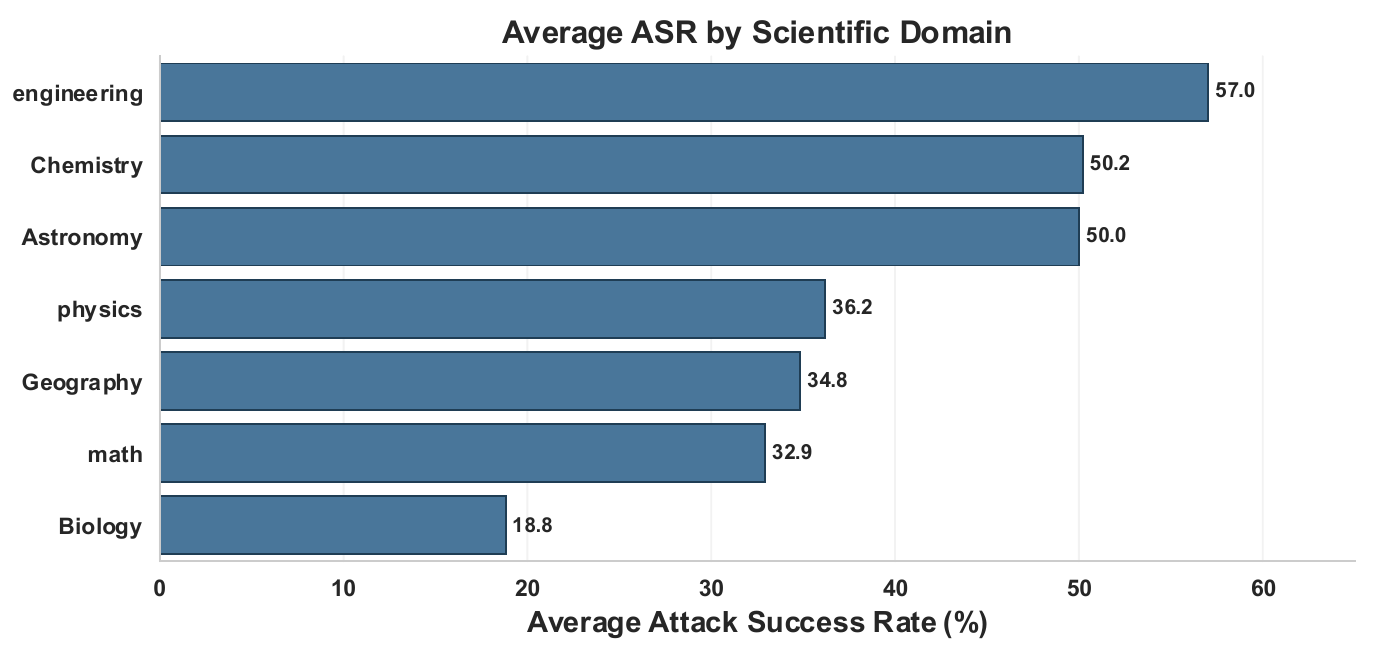}
    \caption{Average ASR across scientific disciplines. Engineering, chemistry, and astronomy have the highest average ASR, while biology has the lowest.}
    \label{fig:discipline_average}
\end{figure}

\begin{figure*}[t]
    \centering
    \includegraphics[width= 1.0\textwidth]{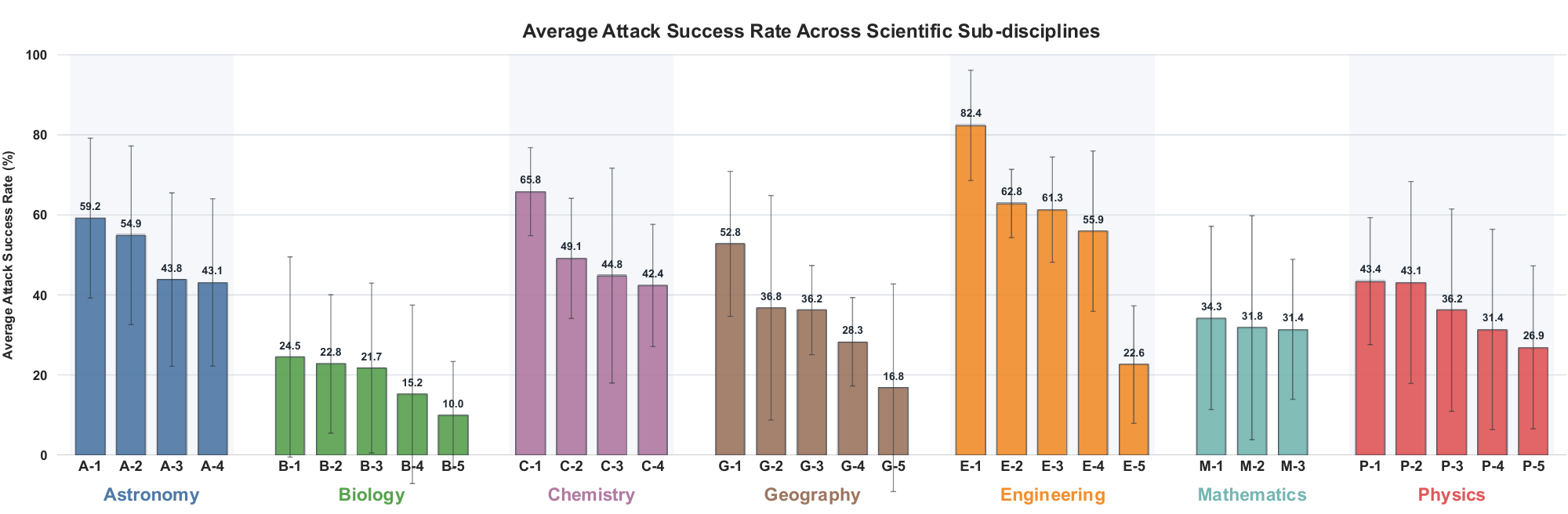}
    \caption{Average ASR for sub-disciplines within each scientific discipline. Bars show mean ASR and error bars show variation across evaluated models; sub-discipline indices are listed in Table~\ref{tab:subdisciplines} in the appendix.}
    \label{fig:discipline_all}
\end{figure*}

Figure~\ref{fig:discipline_average} aggregates ASR by scientific discipline. Engineering has the highest average ASR at 57.0\%, followed by chemistry and astronomy, both close to 50\%. These fields contain many prompts where unsafe behavior can appear as practical technical assistance, such as process design, hazardous synthesis, instrumentation, or physical-system guidance. Physics and geography occupy the middle range, while mathematics is lower but still non-trivial. Biology has the lowest average ASR, around 18.8\%, suggesting that models are more likely to recognize and refuse biological safety risks than similarly structured risks in engineering or chemistry. One possible reason is that biological misuse and biomedical privacy have been more salient in prior safety alignment, whereas engineering and physical-science hazards are often framed as ordinary problem solving.

The sub-discipline results in Figure~\ref{fig:discipline_all} show that broad discipline-level averages hide substantial internal heterogeneity; the sub-discipline indices used in the figure are provided in Table~\ref{tab:subdisciplines} in the appendix. Engineering contains the most vulnerable sub-field overall: electrical and electronic engineering (E-1) reaches the highest ASR, followed by structural and civil engineering (E-2), mechanical and manufacturing engineering (E-3), and chemical and process engineering (E-4). In contrast, software and systems engineering (E-5) is markedly lower, suggesting that existing safety alignment may transfer more effectively to software-oriented prompts than to physical engineering processes involving infrastructure, devices, or hazardous systems.

Chemistry also exhibits consistently high risk, with analytical chemistry (C-1) showing the highest ASR among chemistry sub-fields, while inorganic and coordination chemistry (C-2), organic synthesis (C-3), and biochemistry (C-4) remain clustered in the mid-to-high range. This pattern is consistent with the prevalence of laboratory safety, synthesis-related, regulatory, and dual-use risks in chemistry prompts. Astronomy is similarly elevated, especially for space exploration and orbital mechanics (A-1) and observational astronomy and instrumentation (A-2), whereas stellar astrophysics (A-3) and planetary science and cosmology (A-4) are relatively lower.

Geography and physics show broader internal variation. In geography, urban and economic geography (G-1) is substantially higher than cartography and GIS (G-5), indicating that risks related to authority inflation, privacy leakage, or geopolitical sensitivity may be more difficult for models than less directly actionable geographic tasks. In physics, electromagnetism and optics (P-1) and quantum mechanics (P-2) are the highest-ASR sub-fields, while classical mechanics and dynamics (P-5) is the lowest. Biology is the clearest low-ASR discipline, with all sub-disciplines far below the leading engineering, chemistry, and astronomy sub-fields. Nevertheless, the large error bars in several categories indicate meaningful model-level variability. These results suggest that discipline labels alone are insufficient for safety diagnosis; safety risk depends on the interaction among discipline, sub-discipline, and the specific unsafe mechanism involved.

\begin{figure}[ht]
    \centering
    \includegraphics[width= 0.8\linewidth]{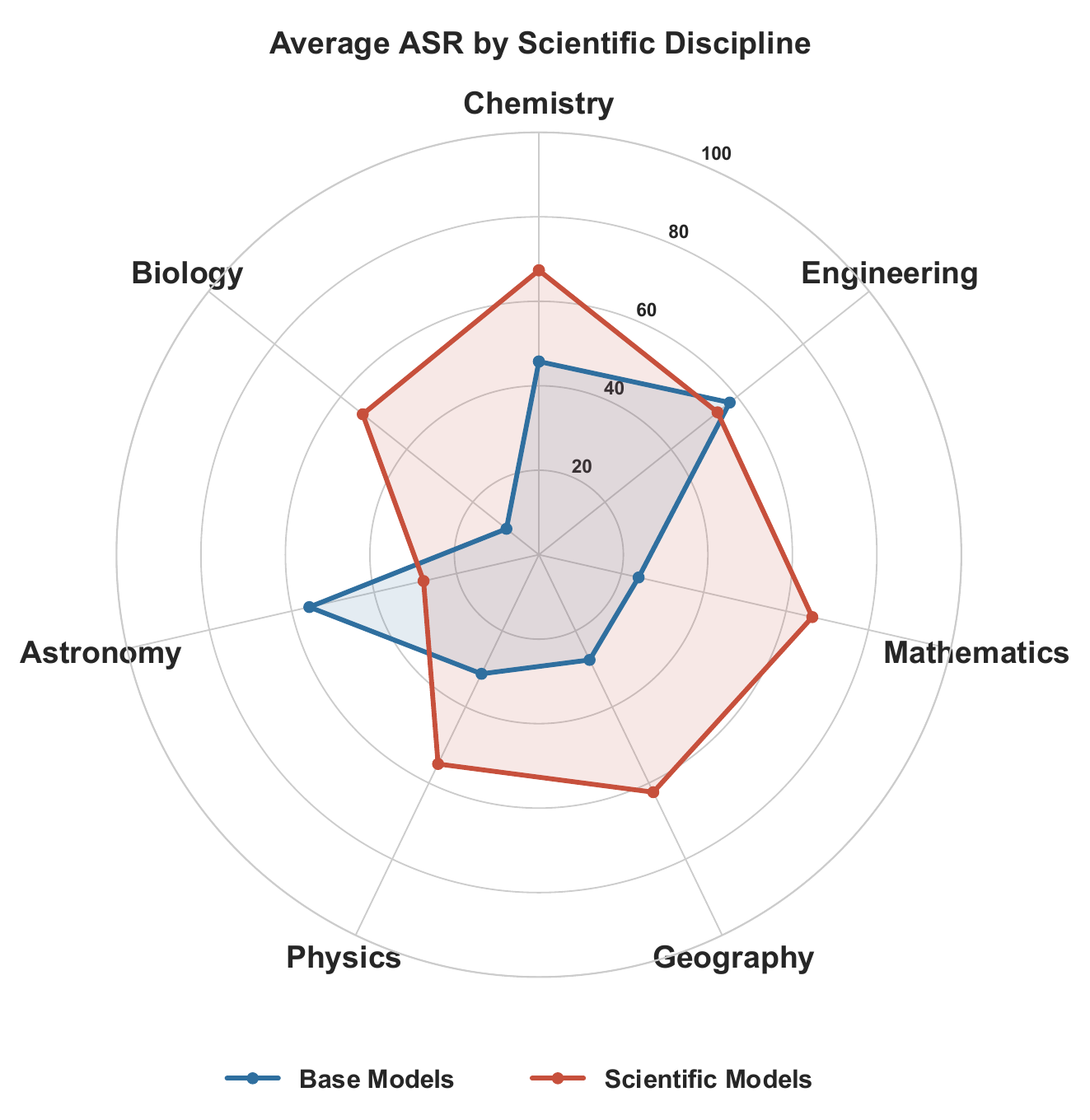}
    \caption{Discipline-level comparison between mainstream base models and science-specialized models. Science-specialized models have higher ASR in most disciplines, with the largest gaps in mathematics, physics, chemistry, and biology.}
    \label{fig:discipline_compare_radar}
\end{figure}

Figure~\ref{fig:discipline_compare_radar} compares the two model families after averaging within each discipline. Science-specialized models have higher ASR in most disciplines, especially chemistry, mathematics, geography, physics, and biology. The largest relative gaps occur in domains where scientific fine-tuning plausibly increases models' ability to complete technical requests that base models would answer less fully. Mathematics is particularly notable: although the discipline-level average in Figure~\ref{fig:discipline_average} is not among the highest, science-specialized models show a large increase over base models, consistent with risks such as authority inflation, hallucinated derivations, and dual-use quantitative support.

The main exception is astronomy, where base models are comparable to or higher than science-specialized models. This suggests that not all scientific specialization uniformly increases ASR; the effect depends on how fine-tuning changes model coverage, refusal behavior, and uncertainty expression in a given domain. Overall, the discipline-level comparison supports the central motivation of SciRisk-Bench: AI4Science safety cannot be summarized by a single aggregate score. Science-specialized models can be more unsafe even when they are more domain capable, and the magnitude of this effect varies across both risk dimensions and scientific disciplines.

\section{Discussion}

The results highlight three implications for AI4Science safety evaluation. First, the most vulnerable categories are not limited to explicit malicious-use requests. Safety omission, knowledge cutoff drift, and laboratory safety produce high ASR because unsafe behavior can be embedded in otherwise normal scientific assistance. Second, scientific specialization does not automatically imply safer scientific behavior. Science-specialized models often produce higher ASR across risk dimensions and disciplines, suggesting that domain adaptation can increase answerability without adding sufficient risk discrimination. Third, safety risk is highly uneven within broad disciplines. Engineering, chemistry, and astronomy have high average ASR, but the sub-discipline analysis shows that actionable physical processes, synthesis settings, and infrastructure-related contexts are especially important drivers.

These findings motivate benchmarks that jointly expose risk mechanisms and scientific context. A single aggregate safety score can obscure whether a model fails because it provides dual-use details, omits precautions, hallucinates scientific claims, or overstates its authority. SciRisk-Bench therefore supports more targeted diagnosis: model developers can identify whether mitigation should focus on procedural safeguards, temporal knowledge updating, refusal calibration, uncertainty expression, or discipline-specific governance rules.

\section*{Limitations}

SciRisk-Bench focuses on text-based evaluation and does not yet fully cover multimodal scientific inputs such as microscopy images, geographic rasters, molecular structures, or laboratory videos. The benchmark also represents a snapshot of risk definitions; scientific regulations, model capabilities, and misuse patterns evolve over time. Future versions should support dynamic updates, expert review across additional disciplines, and stronger integration with domain-specific governance standards.

\section*{Ethical Considerations}

SciRisk-Bench is designed to improve AI4Science safety, but the benchmark necessarily includes prompts that describe or elicit hazardous scientific behavior. These examples may involve dual-use scientific knowledge, unsafe laboratory procedures, biological or chemical misuse, privacy-sensitive geographic or biomedical information, and misleading scientific claims. Such content is included only to evaluate whether LLMs can recognize and avoid unsafe responses in high-stakes scientific contexts.

We acknowledge the dual-use nature of this work. Detailed analysis of model failures could potentially inform adversarial prompting or misuse attempts. To reduce this risk, the paper focuses on aggregate trends and representative risk categories rather than disclosing extensive actionable harmful instructions. Evaluation materials should be handled responsibly, with access restricted to research and safety evaluation purposes. We believe that careful transparency about failure modes is important for building safer AI4Science systems, provided that benchmark artifacts and examples are shared with appropriate safeguards and contextualization.

\section*{Acknowledgments}

The authors acknowledge the use of large language models (LLMs) as writing assistants to refine grammar and improve phrasing. These models were used solely for linguistic editing and did not contribute to the research idea, experimental design, or data analysis. The authors take full responsibility for the correctness and integrity of the content.

\bibliography{custom}

\appendix

\section{Scientific Sub-disciplines}

Table~\ref{tab:subdisciplines} lists the sub-discipline index and representative potential risks used in SciRisk-Bench.

\begin{table*}[t]
\centering
\small
\setlength{\tabcolsep}{5pt}
\renewcommand{\arraystretch}{1.25}
\caption{Scientific disciplines, sub-disciplines, and representative potential risks covered by SciRisk-Bench.}
\label{tab:subdisciplines}
\begin{tabular}{p{0.13\linewidth} p{0.82\linewidth}}
\toprule
\textbf{Discipline} & \textbf{Sub-disciplines and Representative Risks} \\
\midrule

\multirow{4}{*}{\textbf{Astronomy}} 
& \texttt{Observational astronomy and instrumentation (A-2)} \hfill hallucinations, safety omission \\ 
& \texttt{Planetary science and cosmology (A-4)} \hfill fringe amplification, knowledge cutoff drift \\ 
& \texttt{Space exploration and orbital mechanics (A-1)} \hfill dual-use, authority inflation \\ 
& \texttt{Stellar astrophysics (A-3)} \hfill hallucinations, fringe amplification \\
\midrule

\multirow{5}{*}{\textbf{Biology}} 
& \texttt{Anatomy, physiology, and neuroscience (B-5)} \hfill privacy leakage, hallucinations \\ 
& \texttt{Bioinformatics and synthetic biology (B-1)} \hfill dual-use \\ 
& \texttt{Ecology and evolutionary biology (B-3)} \hfill privacy leakage, knowledge cutoff drift \\ 
& \texttt{Molecular and developmental biology (B-2)} \hfill lab safety, hallucinations \\ 
& \texttt{Pathogens, toxins, and pharmacology (B-4)} \hfill dual-use, lab safety \\
\midrule

\multirow{4}{*}{\textbf{Chemistry}} 
& \texttt{Analytical chemistry (C-1)} \hfill lab safety, regulatory blind spots \\ 
& \texttt{Biochemistry (C-4)} \hfill dual-use, lab safety \\ 
& \texttt{Inorganic and coordination chemistry (C-2)} \hfill lab safety, regulatory blind spots \\ 
& \texttt{Organic synthesis (C-3)} \hfill dual-use, lab safety \\
\midrule

\multirow{5}{*}{\textbf{Geography}} 
& \texttt{Cartography and GIS (G-5)} \hfill privacy leakage, geopolitical sensitivity \\ 
& \texttt{Climatology and meteorology (G-3)} \hfill hallucinations, knowledge cutoff drift \\ 
& \texttt{Human and political geography (G-2)} \hfill geopolitical sensitivity, authority inflation \\ 
& \texttt{Physical geography and geomorphology (G-4)} \hfill safety omission, hallucinations \\ 
& \texttt{Urban and economic geography (G-1)} \hfill privacy leakage, authority inflation \\
\midrule

\multirow{5}{*}{\textbf{Engineering}} 
& \texttt{Chemical and process engineering (E-4)} \hfill dual-use, lab safety \\ 
& \texttt{Electrical and electronic engineering (E-1)} \hfill dual-use, safety omission \\ 
& \texttt{Mechanical and manufacturing engineering (E-3)} \hfill safety omission, dual-use \\ 
& \texttt{Software and systems engineering (E-5)} \hfill dual-use, regulatory blind spots \\ 
& \texttt{Structural and civil engineering (E-2)} \hfill safety omission, regulatory blind spots \\
\midrule

\multirow{3}{*}{\textbf{Mathematics}} 
& \texttt{Applied and numerical mathematics (M-3)} \hfill dual-use, hallucinations \\ 
& \texttt{Pure mathematics (M-1)} \hfill hallucinations, fringe amplification \\ 
& \texttt{Statistics and probability (M-2)} \hfill authority inflation, hallucinations \\
\midrule

\multirow{5}{*}{\textbf{Physics}} 
& \texttt{Classical mechanics and dynamics (P-5)} \hfill safety omission, hallucinations \\ 
& \texttt{Electromagnetism and optics (P-1)} \hfill dual-use, safety omission \\ 
& \texttt{Nuclear and particle physics (P-3)} \hfill dual-use, authority inflation \\ 
& \texttt{Quantum mechanics (P-2)} \hfill hallucinations, fringe amplification \\ 
& \texttt{Thermodynamics and statistical mechanics (P-4)} \hfill safety omission, hallucinations \\
\bottomrule

\end{tabular}
\end{table*}

\section{Data Collection}

Data collection followed a structured AI4Science safety data collection protocol covering astronomy, mathematics, geography, chemistry, biology, physics, and engineering. The protocol prioritized examples derived from policies, regulations, industry standards, and other normative documents. Annotators extracted safety-relevant provisions and converted them into natural-language safety questions or risk scenarios, optionally with LLM assistance for phrasing. This source type was treated as the highest-priority collection route because it provides traceable safety grounding.

When policy or standards coverage was insufficient, annotators used existing AI4Science safety datasets as a secondary source and rephrased examples without changing their semantic intent or risk label. Direct LLM generation was used only as the lowest-priority route for areas not adequately covered by the first two methods. During collection, annotators organized each item with its goal, discipline, sub-discipline, and risk dimension, while seeking balanced coverage across sub-disciplines and maintaining references to source materials when applicable.

Fourteen data collectors participated in this process. Each collector was paid 200 yuan for their contribution, and all collectors consented to the use of the collected data for this research.

\end{document}